\documentclass[11pt]{article} % For LaTeX2e
\usepackage{acl}
\usepackage{xcolor}
\usepackage{hyperref}
\usepackage{url}
\usepackage{verbatim}
\usepackage{multirow}
\usepackage{enumitem}
\setlist[itemize]{leftmargin=0.5cm}

\usepackage{latexsym}
\usepackage{booktabs} % for professional tables
\usepackage{makecell}

\usepackage{subfig}

\usepackage[T1]{fontenc}
\usepackage[utf8]{inputenc}

\usepackage{microtype}

\usepackage[cmex10]{amsmath}
\usepackage{amsfonts, amsthm, amssymb}
\usepackage{pifont}% http://ctan.org/pkg/pifont

\definecolor{olivegreen}{RGB}{85, 107, 47}

\usepackage{algorithm}
\usepackage[noend]{algpseudocode}
\usepackage{caption}

\usepackage{tikz}
\usetikzlibrary{arrows, decorations.pathreplacing, decorations.pathmorphing,backgrounds,positioning,fit,petri}
\usepackage[retainorgcmds]{IEEEtrantools}
\newcommand{\be}{\begin{IEEEeqnarray*}{rCl}}
\newcommand{\ee}{\end{IEEEeqnarray*}}

\newcommand{\ben}{\begin{IEEEeqnarray}{rCl}}
\newcommand{\een}{\end{IEEEeqnarray}}

\usepackage{tcolorbox}
\newtcbox{\mybox}[1][blue]{on line,
arc=0pt,outer arc=0pt,colback=#1!10!white,colframe=#1!50!black,
boxsep=0pt,left=2pt,right=2pt,top=2pt,bottom=2pt,
boxrule=0pt,bottomrule=0pt,toprule=0pt}

\title{Skywork-MoE: A Deep Dive into Training Techniques for Mixture-of-Experts Language Models}
\author{\small {Tianwen Wei, Bo Zhu, Liang Zhao, Cheng Cheng, Biye Li, Weiwei L\"u, Peng Cheng} \\
\small {Jianhao Zhang, Xiaoyu Zhang, Liang Zeng, Xiaokun Wang, Yutuan Ma} \\
\small {Rui Hu, Shuicheng Yan, Han Fang, Yahui Zhou}\thanks{\quad Email: \texttt{\{forename\}.\{surname\}@kunlun-inc.com}} \\
\quad \\
Skywork Team, Kunlun Inc.
}

\begin{document}
\maketitle
\begin{abstract}
In this technical report, we introduce the training methodologies implemented in the development of Skywork-MoE, a high-performance mixture-of-experts (MoE) large language model (LLM) with 146 billion parameters and 16 experts. It is initialized from the pre-existing dense checkpoints of our Skywork-13B model. We explore the comparative effectiveness of upcycling versus training from scratch initializations. Our findings suggest that the choice between these two approaches should consider both the performance of the existing dense checkpoints and the MoE training budget. 
%Practical recommendations are provided on when one method may be preferable over the other.
We highlight two innovative techniques: gating logit normalization, which improves expert diversification, and adaptive auxiliary loss coefficients, allowing for layer-specific adjustment of auxiliary loss coefficients. Our experimental results validate the effectiveness of these methods.
Leveraging these techniques and insights, we trained our upcycled Skywork-MoE on a condensed subset of our SkyPile corpus. The evaluation results demonstrate that our model delivers strong performance across a wide range of benchmarks.
\end{abstract}

\section{Introduction}
Recent advancements in the field of artificial intelligence have seen large language models (LLMs) \cite{instruct_gpt, gpt4_report, gpt4_sparks, claude3, llama2, llama3modelcard, gemini_new, deepseekmoev2} revolutionize numerous branches of natural language processing (NLP), encompassing tasks from machine translation to automated summarization. However, the computational demands and associated costs of training and deploying state-of-the-art dense LLMs pose significant challenges, particularly at the scale of tens or hundreds of billions of parameters. In response to these challenges, sparse models, such as Mixture-of-Experts (MoE), have gained prominence \cite{switch_transformer, gshard, glam, deepseekmoev1, deepseekmoev2}. These models offer a more economically viable alternative by distributing computation across various specialized sub-models or ``experts'', potentially matching or even surpassing the performance of their dense counterparts with a fraction of the resource requirements \cite{artetxe2022efficient, deepspeedmoe, moe_scaling_law}.

In light of these developments, this technical report introduces Skywork-MoE, a high-performance MoE large language model with 146 billion parameters and 16 experts. This model leverages the foundational architecture of our previously developed Skywork-13B model \cite{wei2023skywork}, utilizing its dense checkpoints as the initial setup \cite{upcycling}. We conduct experimental analysis on relative benefits of two pivotal strategies in LLM development: upcycling from existing dense models versus initiating training from scratch. Through rigorous evaluation, we provide nuanced insights into how the initial conditions and training budgets influence the effectiveness of these approaches, offering practical guidance on their application. Skywork-MoE embodies the forefront of MoE research by incorporating two novel training techniques: gating logit normalization and adaptive auxiliary loss coefficients. The former aims to enhance the diversification among the experts, while the latter facilitates the tailored adjustment of auxiliary loss coefficients at different layers of the model. Moreover, the training of Skywork-MoE was conducted on a condensed subset of the SkyPile corpus \cite{wei2023skywork}, with subsequent evaluations demonstrating its robust performance across a diverse array of benchmarks. This report aims to detail these innovations and findings, setting a new benchmark for the efficiency and efficacy of MoE models in large-scale language processing tasks.

\section{Preliminaries}
Skywork-MoE follows the previous work of Switch Transformer \cite{switch_transformer}, which implement the idea of MoE \cite{moe1991, moe2014, shazeer2017outrageously} with transformer architecture \cite{VASWANI}.
\subsection{MoE for Transformers}
In a standard transformer, each layer processes inputs through self-attention mechanisms followed by feed-forward neural networks (FFNs) \cite{VASWANI}. The transformer processes every token of the input sequence through the same pathways (i.e., every parameter in the model is active for every input).

In contrast, the MoE architecture modifies the typical transformer by replacing some or all of the FFNs with a mixture-of-experts, where each expert is itself a small FFNs, and the MoE layer houses multiple such experts. The MoE layer increases the capacity of transformer models while maintaining computational efficiency by selectively activating some of the expert networks for each input token. The selection of experts is performed by a gating mechanism, allowing the model to dynamically route tokens to the most relevant experts.

The gating mechanism in consists of a softmax layer that computes a probability distribution over the available experts for each token. The gate output $g$ for the $i$-th token with embedding $x_i$ is given by:
\begin{equation} \label{gating_layer}
\text{softmax}(Wx_i + b) = (g_{i1},\ldots, g_{in})^T
\end{equation}
where $W$ is the gating weight matrix, $b$ is the gating bias vector, ${g}_{ij}$ is the gating probability of the $i$-th token being assigned to the $j$-th expert and $n$ is the total number of experts. The $k$ experts with the highest probability are then selected to process the token, which is also known as top-$k$ routing. Conventionally one chooses $k=1$ or $k=2$. In this work, we always assume using top-$2$ routing of experts.

Let's denote the set of selected experts for the $i$-th token as $\mathcal{E}_i$.
Each selected expert $j \in \mathcal{E}_i$ processes the token embedding ${x}_i$ and generates an output $\text{Expert}_j(x_i)$.
The outputs from the $k$ selected experts are then linearly combined according to the corresponding gating probabilities: 
\begin{equation} \label{output_moe_layer}
y_i = \frac{1}{s_i} \sum_{j \in \mathcal{E}_i}g_{ij} \cdot \mathrm{Expert}_j(x_{i}).
\end{equation}
where $s_i = \sum_{j \in \mathcal{E}_i}g_{ij}$.
The combined output $y_i$ is then passed to the next layer of the model.

\subsection{Auxiliary Loss}
To ensure balanced load across experts and prevent a single expert from dominating, Switch Transformer employs an auxiliary loss function that encourages the even distribution of tokens among experts. Let $p_j$ be the proportions of tokens assigned to expert $j$. The load is balanced across experts if $p_j = {k}/{n}$ for all $j=1,\ldots, n$. An naive auxiliary loss $\mathcal{L}_{\text{aux}}$ that directly penalizes the discrepancy between $p_j$ and $k/n$ would be
\ben \label{naive_loss}
\mathcal{L}_{\text{aux}} = \sum_{j=1}^{n} \left(\frac{k}{n} - p_j \right)^2.
\een 
However, as $p_j$ is only a statistic that does not allow for back-propagation, the naive auxiliary loss is not applicable in practice.  As a differentiable surrogate, one can assume that 
\be
p_j\approx k \cdot E[g_j] \approx \frac{k}{T} \sum_{i=1}^{T} {g}_{ij}
\ee
where $T$ is the number of tokens in a batch.
Substituting $p_j$ by $\frac{k}{T} \sum_{i=1}^{T} {g}_{ij}$ in (\ref{naive_loss}), and ignoring the constant $k$, we obtain
\begin{equation} \label{aux_loss}
\mathcal{L}_{\text{aux}} = \sum_{j=1}^{n} \left(\frac{1}{n} - \frac{1}{T} \sum_{i=1}^{T} {g}_{ij}\right)^2,
\end{equation}
which is the actual auxiliary loss that is commonly used in switch transformer training.
By minimizing this loss, the model can effectively learns to balance the load across experts, preventing any single expert from being overloaded or underutilized.

The total loss function $\mathcal{L}_{\text{total}}$ for training the Switch Transformer is a combination of the cross entropy loss $\mathcal{L}_{\text{ce}}$ for the next token prediction task and the auxiliary loss $\mathcal{L}_{\text{aux}}$, weighted by a hyperparameter $\alpha$:

\begin{equation}\label{total_loss}
\mathcal{L}_{\text{total}} = \mathcal{L}_{\text{ce}} + \alpha \mathcal{L}_{\text{aux}}
\end{equation}

By incorporating the MoE layer and the auxiliary loss for load balancing, Switch Transformer enables the efficient scaling of transformer models to billions of parameters while maintaining computational tractability.

\section{Upcycling vs. From Scratch}
We initiate our discussion by exploring the core issue of upcycling versus training from scratch, a critical consideration in the realm of MoE training. We present our initial experimental findings, comparing the advantages and disadvantages of upcycling from dense model checkpoints versus training a MoE model of equivalent size from scratch.

\subsection{Costs and Budgets}
There are two distinct scenarios:
\begin{itemize}
    \item Sunk Cost: The resources already spent on training the dense model are considered a sunk cost. These are not included in the cost calculations for subsequent upcycled MoE training. This scenario typically applies when utilizing pre-trained dense models, such as those available from open-source platforms.
\item Cumulative Cost: The resources used to train the dense model are included in the total training cost for the upcycled MoE. This occurs when resources are deliberately allocated to first train a dense model, which is then used as a starting point for upcycling.
\end{itemize}
Our discussion will primarily focus on the first scenario, as it will later become clear that allocating resources to train a dense model solely for the purpose of MoE initialization is generally suboptimal.

A priori, the decision to upcycle versus train from scratch should consider the performance of the available dense model and the MoE training budget. On the one hand, if the budget is insufficient to train an MoE from scratch to match or exceed the performance of the dense model, training from scratch is trivially not a sensible option. On the other hand, with ample resources (e.g., significantly more than what was used to train the dense model), training an MoE from scratch might yield better outcomes as it avoids the limitations of starting with a group of identical experts, which can hinder diversification.
\subsection{Experiment Results}
In our experiments, we first train a 0.3B dense model for 300B tokens with peak learning rate $3e$-3 gradually decaying to $3e$-4, obtaining a number of intermediate checkpoints. We focus on upcycling the checkpoints that have undergone 100B and 300B tokens of training, which we denote by ``checkpoint-100B'' and ``checkpoint-300B'' respectively. We then train several MoE models having the same architecture of 8 experts, but with different weight initialization scheme (from-scratch/checkpoint-100B/checkpoint-300B) and peak learning rate.
We conduct this training under two different training budgets: 100 billion and 300 billion tokens.

For the experiments under a budget of 100B tokens, we compare the following:
\begin{figure*}[h]
\centering
\includegraphics[width=1.0\textwidth]{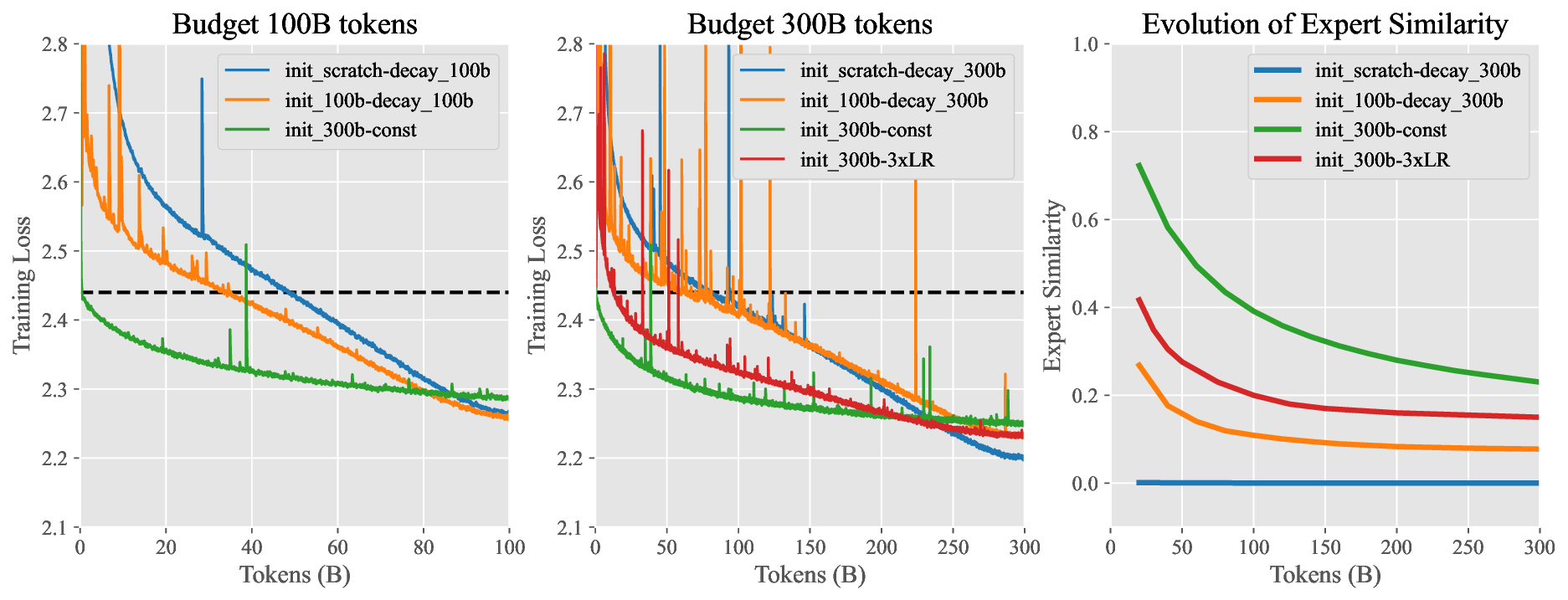} 
\caption{
Training dynamics under different conditions and budgets. Left: Loss curves for MoE training initialized by upcycling and from scratch with 100B token budget. Middle: Similar comparison for a 300B token budget. Right: Evolution of average expert similarity during MoE training with a 300B token budget. The dashed line marks the final loss of a 0.3B dense model at the end of 300B tokens.
}
\label{fig:upcycling_vs_scratch}
\end{figure*}
\begin{itemize}
    \item {\tt init\_scratch-decay\_100b}: From scratch with a peak learning rate of 
$3e$-3 (same as the dense model).
    \item {\tt init\_100b-decay\_100b}: Upcycling from the 100B checkpoint with a peak learning rate of $1.8e$-3.
    \item {\tt init\_300b-const}: Upcycling from the 300B checkpoint with a constant learning rate of $3e$-4.
\end{itemize}
For the  larger 300B tokens budget, we retrain all models with an extended learning rate decay period of 300B tokens. We also train an additional MoE initialized from checkpiont-300B, but with an increased peak learning rate of $1.2e$-3. We denote this model by {\tt init\_300b-3xLR}. Throughout our experiments, we maintain the same minimum learning rate $3e$-4 and decay the learning rate gradually with cosine schedule.

All results are reported in Fig. \ref{fig:upcycling_vs_scratch}. The plot on the left panel indicates that with a moderate budget of 100B tokens, the model trained from scratch achieved similar performance to the model upcycled from checkpoint-100B. Despite starting from a much higher initial loss, both models eventually caught up to and surpassed the performance of the model upcycled from checkpoint-300B. We attribute the poorer performance of the latter to its overly small learning rate of $3e$-4. 
The plot in the middle reveals that with a larger budget of 300B tokens, the model trained from scratch outperforms all of its upcycled counterparts. Among the upcycled models, the one trained with the smallest learning rate again delivers the poorest result, underscoring the critical role of learning rate schedules in training MoE models.
The plot on the right shows the decreasing trend of the average expert similarity during training for the upcycled MoEs, revealing that the process of training an upcycled MoE involves the diversification of experts. Notably, the model with the highest expert similarity exhibits the weakest performance, reinforcing the idea that expert similarity can serve as an effective monitoring metric during MoE training when models are initialized through upcycling. In contrast, throughout the training, the expert similarity for the from-scratch MoE remains at zero, suggesting that a non-uniform expert initialization encourages diversification.

\subsection{Rules of Thumb for Upcycling}
Let us denote by $C$ the cost of training an 0.3B dense model for 300B tokens. Then, for a corresponding MoE moddel, the training costs for 100B and 300B tokens are roughly $\frac{2}{3}C$ and $2C$ respectively\footnote{This estimation is based on our use of top-2 routing in the MoE model, which results in approximately 1.7 times the number of activation parameters compared to the dense model. If we also take into account of the communication overhead associated with expert parallelism, training the MoE model requires roughly twice the GPU hours compared to its dense counterpart for the same number of tokens. }.
Our experiment results state that in our setting with a moderate training budget of $\frac{2}{3}C$, an MoE trained from scratch is able to achieve similar performance to an upcycled one, initialized from dense checkpoints that has undergone pre-training of budget $C$. If, however, the training budget for MoE is $2C$, twice of the training budget of the dense checkpoint, then an MoE trained from scratch performs significantly better than its upcycled counterpart. 

Let us denote by $C_{\textrm{dense}}$ the cost to train the dense model from which one can choose to upcycle from for the MoE training, and by $C_{\textrm{MoE}}$ the training budget for the MoE model itselt.
Our findings suggests the following rule of thumb on whether or not to adopt upcycling when upcycling is possible is given as follows:
\begin{itemize}
\item If $C_{\text{MoE}} \ll C_{\textrm{dense}}$, then one should prefer upcycling over training from scratch to maximally exploit the sunk cost invested in the dense model. 
\item If $C_{\text{MoE}} \geq 2C_{\textrm{dense}}$, then one should stick to the conventional method of training from scratch over upcycling, as the benefit of upcycling from a pre-trained checkpoint cannot compensate for the difficulty of expert diversification due to the uniformity of initialized experts. 
\item If one does not have a pre-trained dense checkpoint to upcycle from, then this corresponds to the case $C_{\text{MoE}} \gg C_{\textrm{dense}}=0$. As a consequence, one should always train the MoE from scratch.
\item When training an upcycled MoE, one should carefully tune the learning rate schedule. Different learning rate schedule may yield different 
\end{itemize}

\section{Training Techniques}
\subsection{Gating Logit Normalization}
One phenomenon that we have frequently observed during the training of MoE models is that its gating layers sometimes tend to yield distributions with high entropy, i.e., the top-$k$ probabilities for the selected experts are only marginally greater than those for the non-selected experts. 
Consequently, the output of the MoE layer is approximated as follows:
\be
y_i \approx \frac{1}{k} \sum_{j \in \mathcal{E}_i} \mathrm{Expert}_j(x_{i}),
\ee
In this scenario, the output is effectively a simple average of the selected expert outputs, rather than a weighted average. This suggests a uniformity among experts, indicating that the gating mechanism fails to discriminate effectively between different experts, which can be detrimental to model performance.
\begin{figure}[h]
\centering
\includegraphics[width=0.45\textwidth]{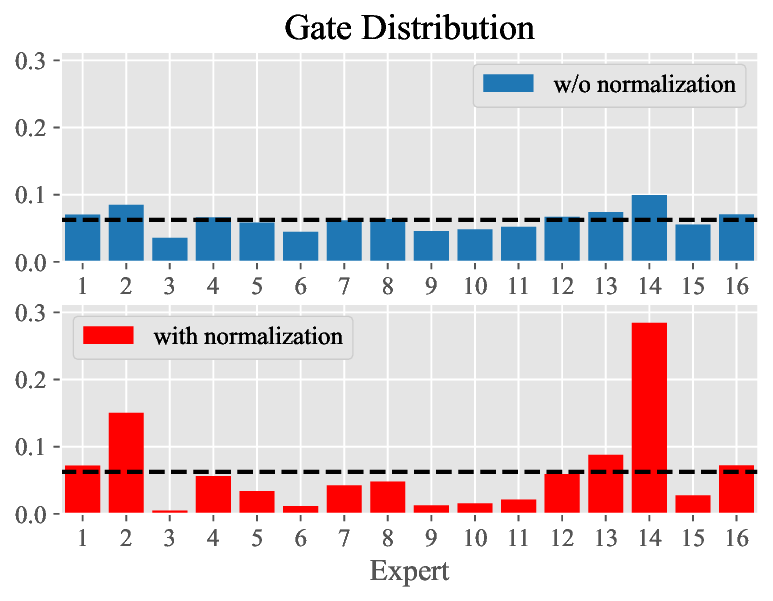} 
\caption{A comparison of gate distribution with and without logit normalization. The black dashed line corresponds to the baseline of uniform probability $1/16$.}
\label{fig:gate_distribution}
\end{figure}

Although the underlying cause of this phenomenon still warrants further investigation, we have identified a straightforward solution. This remedy involves introducing a normalization step prior to the softmax function in the gating layer to ensure a more distinct gate output distribution. Specifically, we propose modifying the gating layer (\ref{gating_layer}) as follows:
\ben
z & = & Wx + b   \nonumber \\
%$\mu, v & \leftarrow & \textrm{avg}(z), \textrm{std}(z)  \\
\tilde{z} & = & \lambda \cdot \frac{z - \mu}{\sigma}  \label{normalization} \\
g & = & \text{softmax}( \tilde{z}), \nonumber
\een
In this revised formulation, the vector $z$ is first normalized by subtracting its mean $\mu$ and dividing by its standard deviation $\sigma$. It is then scaled by a hyper-parameter $\lambda$, resulting in a transformed vector $\tilde{z}$ with zero mean and a standard deviation controlled by $\lambda$. This adjustment ensures that the output vector $\tilde{z}$ is suitably scaled before applying the softmax function.
The parameter $\lambda$ plays the important role of determining the sharpness of the softmax output distribution. Specifically, a higher value of $\lambda$ leads to a sharper, more focused distribution. This sharper gating mechanism is intended to enhance the model's ability to effectively differentiate between the contributions of various experts, thereby potentially improving the overall performance of the MoE model.

\begin{figure*}[h]
\centering
\includegraphics[width=0.9\textwidth]{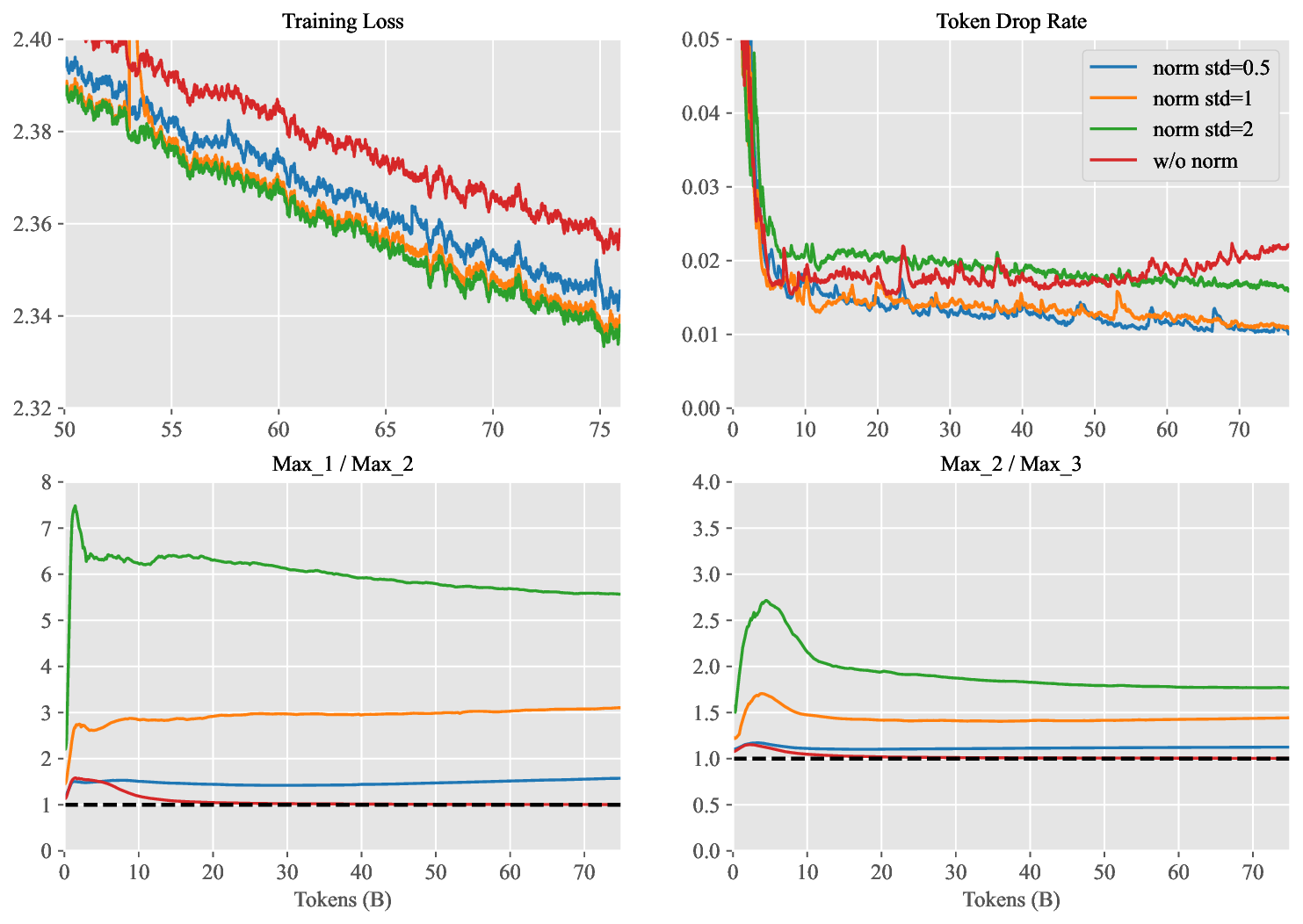} 
\caption{Top left: Training loss curves for MoE models with and without gating normalization, illustrating that gating normalization contributes to a moderate improvement in loss. Top right: Evolution of the token drop rate for each model, showing the regularization effect of gating normalization which helps to reduce token drop during gating.  Lower: Ratios $\mathrm{Max}_1 / \mathrm{Max}_2$ and $\mathrm{Max}_2 / \mathrm{Max}_3$ from the softmax output of the 3rd gating layer throughout training. We observe that a higher \texttt{std} parameter value increases both ratios as expected. For the model trained without gating normalization, both ratios converge to one (indicated by the horizontal dashed line), a condition considered detrimental for the model's performance.
}
\label{fig:router_normalization}
\end{figure*}

To validate our proposed methodology, we conducted a small-scale experiment using an MoE model equipped with 2.5 billion parameters and 16 experts. We compared models trained both with and without gating logit normalization and varied the hyperparameter $\lambda$. 
The results are illustrated in Figure \ref{fig:gate_distribution} and Figure \ref{fig:router_normalization}. 
In Figure \ref{fig:gate_distribution} we show the output distribution of a gate for a model trained with gating logit normalization is significantly sharper than the one trained without. In the upper plots of Figure \ref{fig:router_normalization}, we can see that all models trained with gating logit normalization exhibit significantly lower training losses and token drop rates compared to that without normalization.
Additionally, we analyzed the ratios of $Max_1 / Max_2$ and $Max_2 / Max_3$, where $Max_i$ represents the $i$-th largest probability in the gate output distribution. These ratios are important indicators of the discriminative power of the expert router. A higher $Max_1 / Max_2$ and $Max_2 / Max_3$ ratio suggests a more effective differentiation among experts. As shown in the lower plots of Figure \ref{fig:router_normalization}, increasing $\lambda$ leads to higher ratios, aligning with expectations. However, since the training losses for $\lambda=1$ and $\lambda=2$ are comparably effective, we have chosen to implement $\lambda=1$ in the training of our Skywork-MoE model. 

\subsection{Adaptive Auxiliary Loss Coefficients}
%[Discuss the effect of inappropriate $\lambda$]

The primary purpose of integrating an auxiliary loss (\ref{aux_loss}) is to facilitate a balanced distribution of workload across experts during training. This balance not only ensures effective training for each expert but also fosters diversity among them. The intensity of this load balance regularization is governed by a tunable hyper-parameter, $\alpha$, which is commonly set to either $1e$-2 or $1e$-3 in practical applications.

We present two key observations. Firstly, since each gating layer possesses its independent auxiliary loss, the coefficients corresponding to these losses do not necessarily have to be identical. In that regard, a more explicit form of the total loss (\ref{total_loss}) should be
\be
\mathcal{L}_{\text{total}} = \mathcal{L}_{\text{ce}} + \sum_{l=1}^M\alpha^{(l)} \mathcal{L}_{\text{aux}}^{(l)},
\ee
where $M$ is the total number of MoE layers, and $\mathcal{L}_{\text{aux}}^{(l)}$ and $\alpha^{(l)}$ are auxiliary loss and its coefficient for the $l$-th  MoE layer, respectively. We speculate that there may exist a combination of ``optimal'' coefficient values that is superior to a single fixed global auxiliary loss coefficient applicable to all layers. 

Secondly, if the load is already balanced across the experts during training, then it is advisable to reduce the auxiliary loss coefficients to alleviate the load balance regularization. On the contrary, in scenarios where there is a significant imbalance in load distribution among experts, increasing the coefficients would enforce stricter load balance regularization. The rationale for adjusting these coefficients is primarily to prioritize the optimization of the cross-entropy loss for next-word prediction, while treating load balance regularization as a secondary, potentially counterproductive, goal.

To address this, we propose the method of \emph{Adaptive Auxiliary Loss Coefficients}. This approach involves monitoring the token drop rate, which we use as a measure for expert load balance, for each MoE layer throughout the training process, and adaptively updating the coefficients for subsequent iterations based on the observed token drop rates. The updates to the loss coefficients are designed to be positively correlated with the token drop rates.

More specifically, we define the update mechanism as follows:
\begin{eqnarray}
\hat{\alpha}_{i+1}^{(l)} & = & f(d_i^{(l)}), \\
{\alpha}_{i+1}^{(l)} & = & \beta {\alpha}_{i}^{(l)} + (1-\beta) \hat{\alpha}_{i+1}^{(l)},
\end{eqnarray}
where:
\begin{itemize}
    \item $f$ is an increasing function mapping the current observed token drop rate $d_i^{(l)}$ to an estimated auxiliary loss $\hat{\alpha}_{i+1}^{(l)}$ for the next iteration.
\item ${\alpha}_{i+1}^{(l)}$ represents the moving average of $\hat{\alpha}_{i+1}^{(l)}$, serving as the actual auxiliary loss coefficient for the next iteration. This moving average approach mitigates abrupt changes in regularization intensity.
\item $\beta$, a parameter within the range (0, 1), balances the weight between the existing moving average and the new estimate.
\end{itemize}
In our specific implementation, we define $f(d) = \xi d$ for some $\xi > 0$, with the constraint that $f(d)$ does not exceed a maximum value $c_{\max}$. This results in a piece-wise linear function:
\begin{eqnarray}
f(d) = \left\{
\begin{array}{ll}
\xi d &  \textrm{if } d \leq {\alpha_{\max}}/{\xi}, \\
\alpha_{\max} & \textrm{if } d > {\alpha_{\max}}/{\xi}.
\end{array}
\right.
\end{eqnarray}
The hyper-parameter $\xi$ regulates the sensitivity of the loss coefficients to the token drop rate. During our training of the Skywork MoE model, we set $\xi=1/5$, $\alpha_{\max}=0.01$, and $\beta=0.99$. This configuration effectively maintained both token drop rates and auxiliary loss coefficients at desirable levels.

\begin{figure}[h]
\centering
\includegraphics[width=0.48\textwidth]{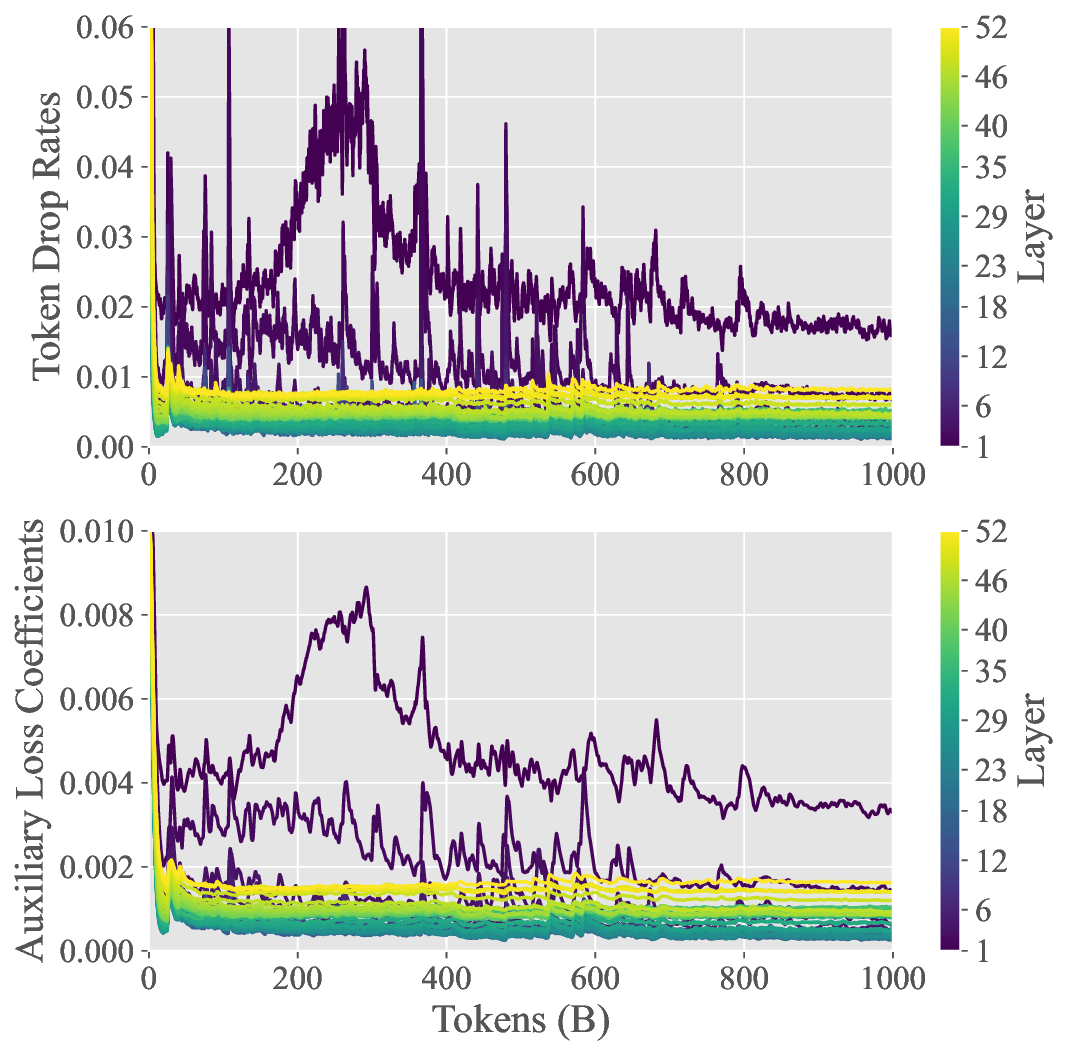} 
\caption{The curves of token drop rate (top) and those of auxiliary loss coefficient (bottom) for all gating layers during the pre-training of our Skywork-MoE. It can be seen that the auxiliary loss coefficients is responsive to the change in token drop rates.
}
\label{fig:drop_rate}
\end{figure}

\begin{table*}[h]
\centering
\resizebox{0.95\textwidth}{!}{%
\begin{tabular}{l|cc|cccccc}
\toprule
& \textbf{\#AP} & \textbf{\#TP} & \textbf{CEVAL} &\textbf{CMMLU} & \textbf{MMLU} & \textbf{GSM8K} & \textbf{MATH} & \textbf{HumanEval} \\
\midrule
%{Yi-34B} & 34& 34 & 81.4 & 83.7 & 76.3 & 76.0 & 15.3 & 26.2 \\
{Deepseek-67B} & 67 & 67 & 66.1 & 70.8 & 71.3 & 63.4 & 18.7 & 42.7 \\
{Qwen1.5-72B} & 72 & 72 & 84.1 & 83.5 & 77.5 & 79.5 & 34.1 & 41.5 \\
{Llama2-70B} & 70 & 70 & - & - & 68.9 & 56.8 & 13.6 & 29.9 \\
{Llama3-70B} & 70 & 70 & - & - & 78.8 & 82.7 & 36.7 & 39.0 \\
\midrule
{Mixtral 8*7B} & 13 & 47 & - & - & 70.6 & 58.4 & 28.4 & 40.2 \\
{Mixtral 8*22B} & 39 & 141 & - & - & 77.8 & 78.6 & 41.8 & 45.1\\
{Grok-1} & 86 & 314 & - & - & 73.0 & 62.9 & 23.9 & 63.2 \\
{DBRX-Instruct} & 36 & 132 & - & - & 73.7 & 66.9 & - & 70.1 \\
{Deepseek-V2} & 21 & 236 & 81.7 & 84.0 & 78.5 & 79.2 & 43.6 & 48.8 \\
\midrule
{Skywork-13B} &13 & 13 & 62.1 & 62.4 & 62.7 & 60.2 & 8.4 & 18.9 \\
{Skywork-MoE} & 22 & 146 & 82.2 & 79.5 & 77.4 & 76.1 & 31.9 & 43.9 \\
\bottomrule
\end{tabular}    
}            
\caption{Evaluation results of Skywork-MoE on popular LLM benchmarks. Results of recent open models are also reported for comparison. The columns titled ``\#AP'' and ``\#TP'' stand for the number of activated parameters and that of total parameters (in billion), respectively.}
\label{tab:evaluation}
\end{table*}

\section{Skywork-MoE}

Skywork-MoE is a massive MoE model with a total of 146 billion parameters and 22 billion activated parameters. It initialized from our in-house pre-trained Skywork-13B \cite{wei2023skywork} dense checkpoint\footnote{The open sourced version of Skywork-13B has been trained for 3.2 trillion tokens. the in-house version has undergone additional pre-training on an extra 2 trillion tokens.}, and is trained with gating logit normalization and adaptive auxiliary loss coefficient.

Skywork-MoE has undergone several stages of training, each characterized by a unique learning rate schedule and composition of training data. The data utilized to train Skywork-MoE consists of a curated subset of our SkyPile corpus \cite{wei2023skywork}, enriched with a significant volume of synthetic data.
Overall, the collective distribution of training data aligns with a ratio of approximately 7:2:1 among English, Chinese, and code data.

To evaluate the performance of Skywork-MoE, we consider the following popular benchmarks:
%CEVAL \cite{ceval} and CMMLU \cite{cmmlu} for knowledge and problem solving in Chinese ,  MMLU \cite{mmlu} for English, GSM8K \cite{gsm8k} and MATH \cite{hendrycks2021} for mathematical reasoning and HumanEval\cite{humaneval} for programming capability. 
To assess the model's knowledge and problem-solving skills in Chinese, we utilized the CEVAL \cite{ceval} and CMMLU \cite{cmmlu} benchmarks. The MMLU \cite{mmlu} benchmark was chosen to evaluate English proficiency. For testing mathematical reasoning, the GSM8K \cite{gsm8k} and MATH \cite{hendrycks2021} datasets were included. Additionally, the model's programming capabilities were assessed using the HumanEval \cite{humaneval} dataset.

We also present benchmark results for recent open-source models of comparable size, encompassing both dense and MoE architectures. Those models include: Deepseek-67B \cite{deepseek67}, Qwen1.5-72B \cite{qwen}, Llama2-70B \cite{llama2}, Llama3-70B \cite{llama3modelcard}, Mixtral 8*7B \cite{mixtral_8_7B}, Mixtral 8*22B \cite{mixtral_8_22B}, DBRX-Instruct \cite{dbrx}, Deepseek-V1 \cite{deepseekmoev1}, Deepseek-V2 \cite{deepseekmoev2}.

The evaluation results are presented in Table \ref{tab:evaluation}. It can be seen that Skywork-MoE achieves strong scores of 82.2 and 79.5 on the CEVAL and CMMLU benchmarks, respectively, surpassing Deepseek-67B, and is closely trailing behind Deepseek-V2. On the MMLU, Skywork-MoE scores 77.4, which is competitive when compared to higher-capacity models like Qwen1.5-72B and slightly lower than Llama3-70B. 
In mathematical related tasks (GSM8K and MATH), Skywork-MoE's scores of 76.1 and 31.9 are notable. It comfortably outperforms Llama2-70B and Mixtral 8*7B and stands close to larger models such as Deepseek-V2 (79.2 and 43.6). This highlights the model's ability to handle complex quantitative and logical reasoning, a challenging area for many language models.
On the HumanEval benchmark, which tests code synthesis capabilities, Skywork-MoE scores 43.9. This is a strong performance, exceeding all dense models in our comparison. It is slightly below Deepseek-V2, suggesting room for improvement in programming-related tasks.
Overall, it is pertinent to conclude that our Skywork-MoE outperforms Deepseek-67B and Llama2-70B, but trails behind Llama3-70B and several larger MoEs such as Mixtral 8*22B and Deepseek-V2.

\section{Conclusion}
In this work we introduced the techniques and insights we gained behind the development of the Skywork-MoE model. Our comparative analysis of upcycling pre-existing models versus training from scratch provides insights and guidelines into the initization decisions required for MoE model development. This understanding allows for more informed and effective planning and allocation of resources in large-scale MoE training projects.
We introduced gating logit normalization and adaptive auxiliary loss coefficients, two techniques that have notably enhanced expert diversification and provided a flexible framework for adjusting auxiliary losses, respectively. Based on these findings, we trained Skywork-MoE, an open-source MoE upcycled from previous Skywork-13B checkpoint. Its strong performance validates the effectiveness of our approach.

\bibliographystyle{acl_natbib}
\bibliography{llm}
\appendix
\section{Skywork-MoE Architecture}
As Skywork-MoE is upcycled from Skywork-13B, the MoE inherits most of the network configuration of the latter model, which is of Llama-like \cite{llama, llama2} architecture featuring Rotary Positional Embedding (RoPE) \cite{rope}, RMSNorm \cite{rmsnorm} and SwiGLU activation function \citep{swiglu}. Other details on Skywork-MoE is given in Table \ref{table:architecture}.
\begin{table}[ht]
\renewcommand{\arraystretch}{1.0} % Adjust the row height
\centering
%\resizebox{0.45\textwidth}{!}{%
\begin{tabular}{l|c}
\toprule
 & Skywork-MoE \\
\midrule
Vocab. Size  & 65,536 \\
Hidden Dim. & 4,608\\
FFN Dim.    & 12,288\\
Head Dim.   & 128 \\
Num. Heads  & 36\\
Num. Layers & 52 \\
\midrule
Num. Total Experts & 16 \\
Num. Routed Experts & 2 \\
MoE Layer Frequency & 1 \\
\midrule 
Native Seq. Len. &  8192 \\
\bottomrule
\end{tabular}
%}
\caption{Details on Skywork-MoE architecture.}
\label{table:architecture}
\end{table}

\section{Infrastructure}
The Skywork-MoE model leverages our internally developed training framework, Skywork-Megatron, which is built on the Megatron-LM \cite{megatronlm, megatron2021} 23.06 branch. Within this framework, we have implemented a custom MoE architecture that includes gating layer, expert layer, and a tailored distributed parallel strategy.

\subsection{Expert Data Parallel (EDP)}

\begin{figure}[h]
\centering
\includegraphics[width=0.5\textwidth]{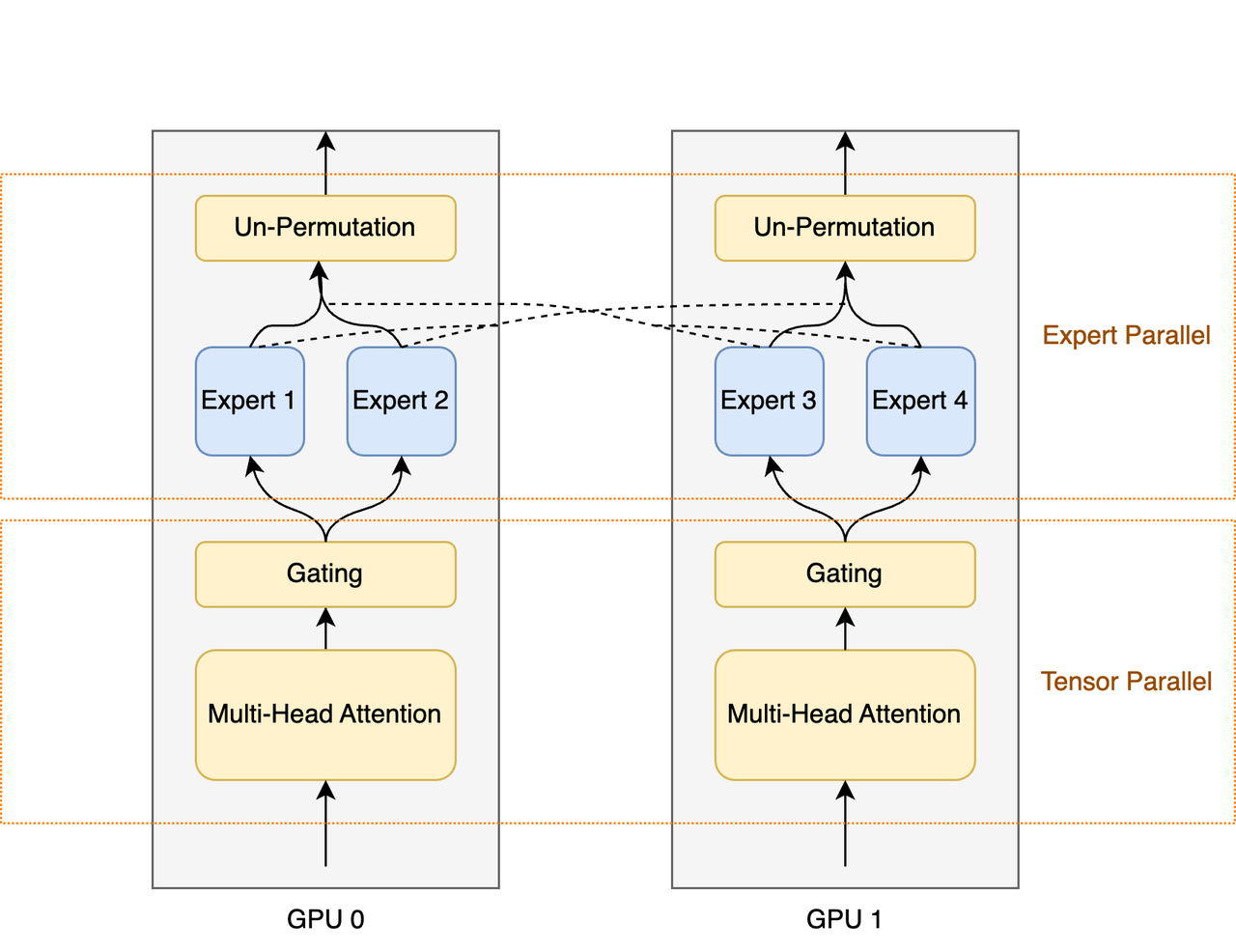} 
\caption{Illustration of Expert Data Parallism (EDP). In EDP, the attention part runs as Tensor Parallelism, while the FFN part runs as Expert Parallelism.}
\label{fig:EDP}
\end{figure}

We introduces a unique parallelization strategy named \emph{Expert Data Parallelism} (EDP). 
Existing parallelism strategies for MoE training in Megatron-LM Core 0.6.0 include Expert Parallelism (EP) and Expert Tensor Parallelism (ETP).
\begin{itemize}
\item EP is characterized by $\textrm{Size}_{EP} = \textrm{Size}_{DP} * \textrm{Size}_{TP}$. As EP does not support further split of single expert, there is also a constraint that $\textrm{Size}_{EP}$ cannot exceed the total number of experts. Consequently, with EP the number of GPUs that can be used to train the MoE is bounded by a multiple of the number of experts.
\item ETP is characterized by $\textrm{Size}_{EP} = \textrm{Size}_{DP}$. As ETP  allows splitting one expert onto multiple GPUs ($ \textrm{Size}_{TP}$), it supports larger cluster size than that of EP. The downside is that ETP has a larger communication overhead fom AlltoAll operation between experts, which my increases rapidly with $\textrm{Size}_{TP}$.
\end{itemize}
Our EDP is defined by $\textrm{Size}_{EP} = \textrm{Size}_{TP}$. This approach is particularly effective for models with a moderate number of experts (e.g., no greater than 64), optimizing the AllToAll communication during the routing of tokens by the gating layer. In the EDP configuration (see Figure \ref{fig:EDP} for an illustration), the same data traverses both the TP Group in the attention layer and the EP Group in the expert layer. The device mesh configuration for Attention and Expert weights is represented as $[\textrm{Size}_{PP}, \textrm{Size}_{DP}, \textrm{Size}_{TP}]$ and $[\textrm{Size}_{PP}, \textrm{Size}_{DP}, \textrm{Size}_{EP}]$, respectively.

%![Diagram of EDP Configuration][]
\subsection{Unbalanced Pipeline Parallellism}
\begin{figure*}[h]
\centering
\includegraphics[width=\textwidth]{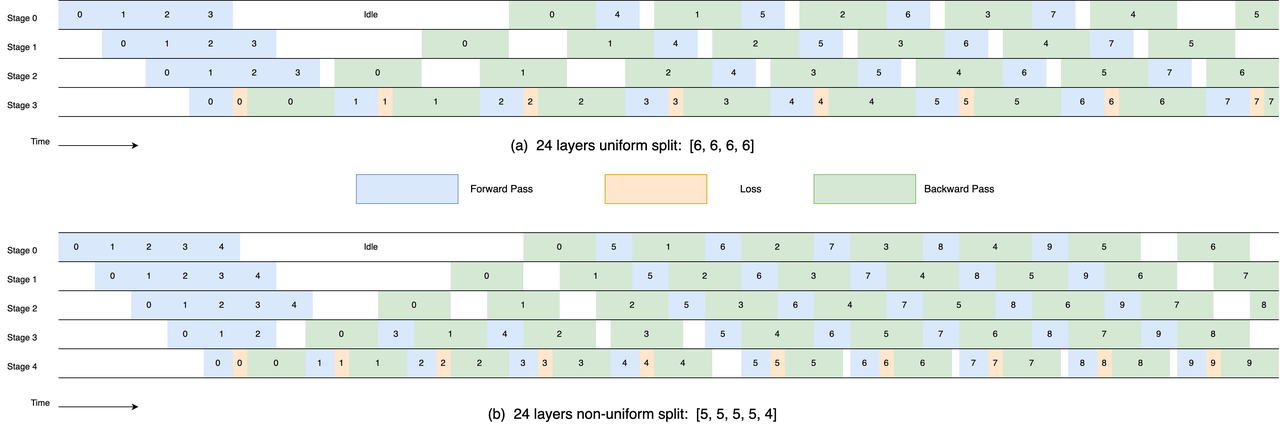} 
\caption{Comparison of bubble time between uniform and non-uniform split pipeline parallelism (PP) in a 24-layer transformer network. (a) Uniformly split into four PP stages, each containing six layers, resulting in significant bubble formation due to the computational demands of loss calculation. (b) Non-uniformly split into five PP stages configured as [5, 5, 5, 5, 4], with the final stage containing one fewer layer, achieving better load balance across stages.}
\label{fig:PP}
\end{figure*}

The Skywork-MoE model employs a custom approach to Pipeline Parallelism (PP) and gradient recomputation to achieve better load balancing across both GPU computation and memory usage in various pipeline stages. Standard pipeline parallel implementations often suffer from computational bottlenecks, particularly in the last stage due to the loss calculation. In Figure \ref{fig:PP} we present an example of a model with 24 layers. In this example, 
adjusting the segmentation of transformer layers from a uniform $[6, 6, 6, 6]$ to $[5, 5, 5, 5, 4]$ reduces pipeline bubble time by up to 10\%, enhancing overall computational efficiency.
%![Pipeline Stage Segmentation][]
Similarly, gradient recomputation (via checkpointing) is adapted differently across the stages. With large differences in buffer sizes across the stages, configuring varied recomputation layer numbers for each stage helps in balancing memory utilization and computational overhead effectively.

\subsection{Training Efficiency}
The training of the Skywork-MoE model is conducted on a cluster comprising 192 NVIDIA-HGX-A800 nodes, totaling 1536 A800-80G SXM GPUs. Each node is connected through a high-speed 400 GB/s NVLink for intra-node and an 800 Gb/s RoCE network for inter-node communications. The model utilizes 12-way pipeline parallelism, 4-way tensor-expert parallelism (via EDP), and 32-way data parallelism with ZeRO-1 optimization \cite{zero}. To further enhance training performance, we have implemented features such as communication reduction related to expert parallelism, kernel fusion, and overlapping communication with computation. 

Ultimately, the training of Skywork-MoE achieves 38\% Model Floating-point Utilization (MFU) on the cluster and a throughput of 690 tokens per GPU per second.

\section{Negative Results}
\subsection{Scaling Expert Learning Rate}
In MoE training with top-$k$ routing, each input token is assigned to $k$ experts. If the expert loads are roughly balanced, then in a forward pass each expert is expected to receive a proportion of $k/n$ of all input tokens. This means that the effective training batch size for the MoE layers is merely $k/n$ of the nominal training batch size. As smaller effective batch size leads to more noised gradient estimate, one may hypothesize that to compensate this it is preferrable to scale the learning rate of the MoE layer by a factor of either $k/n$ (linear scaling) or $\sqrt{k/n}$ (squre root scaling).

In order to test the validity of such treatment, we have experimented with a small MoE model featuring 32 experts and a total of 1.8 billion parameters, utilizing top-2 routing with 150 million activated parameters. Under this setting, the effective batch size for the MoE layers is 16 times smaller than the nominal batch size. With the square root scaling, the learning rate for the MoE layer should be scaled by $1/\sqrt{16}=0.25$.

We have experimented with three different learning rate setting:
\begin{itemize}
    \item {Baseline}: a global peak learning rate of $6e$-3 for all component of the network;
    \item {Expert lr $\times0.25$}: the peak learning rate is set to be $1.5e$-3 for MoE layers and $6e$-3 for non-MoE layers;
    \item {Baseline lr $\times0.25$}: a global peak learning rate of $1.5e$-3 for all component of the network.
\end{itemize}
All models were first trained from scratch for 300 billion tokens, and learning rate linearly decreasing to $10\%$ of its peak value. We then continued the training for another 10B tokens, during which the learning rate is swiftly decayed from the its final value in the previous stage to zero.

\begin{figure}[h]
\centering
\includegraphics[width=0.48\textwidth]{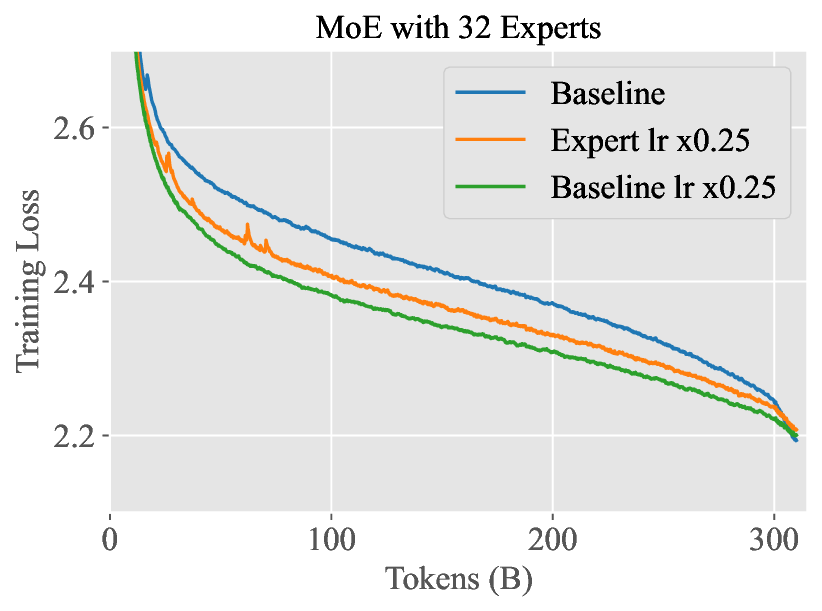} 
\caption{
Comparison of Expert vs. Global Learning Rate Scaling. This graph illustrates the noticeable differences in training loss at 300 billion tokens, attributable to variations in their terminal learning rates. However, by 310 billion tokens, when the learning rate reaches zero, the training curves of all three models converge, demonstrating similar performance outcomes.
}
\label{fig:expert_lr}
\end{figure}

The experiment result is depicted in Fig. \ref{fig:expert_lr}. We see that at the end of the first stage of training, the baseline models with and without global learning rate scaling exhibits the best and poorest performance respectively, and the model with expert learning rate scaling is somewhere in-between. We attribute this performance gap to the \emph{difference} of their respective final learning rate. This can be evidenced by the fact that with merely 10B additional training, where the learning rates for all models had declined to zero, only minor differences in training loss remained, with the baseline model marginally outperforming the others.

%In either case, we deem that there is no need to adjust the learning rate for MoE layers, despite the theoretical support. One possible explanation is that with the configuration of 32 experts, the parameters in the MoE layers already account for $97\%$ of all model parameters. Therefore, scaling the learning rate for MoE layers is almost equivalent to scaling the learning rate globally.

Despite theoretical justifications for adjusting the learning rate for MoE layers, our findings suggest that such modifications may be unnecessary. We note that in our configuration of 32 experts the parameters within the MoE layers constitute approximately 97\% of the total model parameters, where the latter figure mainly depends on the number of experts and is agnostic to the model scale. Consequently, adjusting the learning rate specifically for the MoE layers effectively equates to a global scaling of the learning rate across the entire network. This overlap in parameter distribution implies that targeted adjustments to the MoE layer's learning rate might not yield distinct outcomes from global adjustments.
\begin{figure}[h]
\centering
\includegraphics[width=0.48\textwidth]{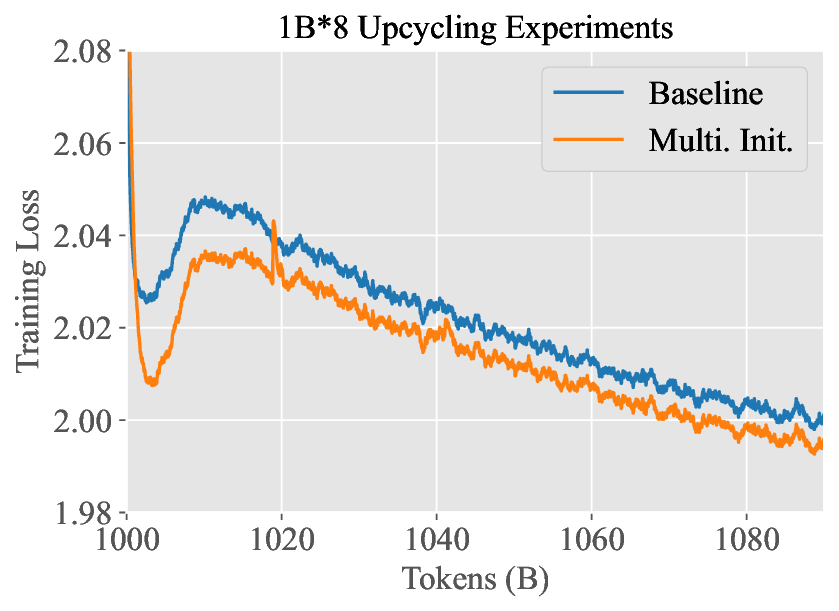} 
\caption{Comparison of training loss for MoE models: conventional upcycling (baseline) versus specialization training (Multi. Init.). Both models underwent training over 100 billion tokens.
% Note the initial 10 billion tokens serve as a warm-up phase, which results in an increased loss.
}
\label{fig:multi_init}
\end{figure}

\subsection{Expert Specialization Training for Upcycling}
Conventional sparse upcycling methods involve initializing MoE weights from a single dense model checkpoint, where the weights in the Feed-Forward Network (FFN) layers of the dense model are replicated \(n\) times, creating an MoE model with \(n\) \emph{identical} experts in each MoE layer. It is reasonable to hypothesize that this method of initializing MoE models with identical experts could impede the diversification of the experts, potentially leading to suboptimal performance.

To investigate this, we explored a method which we refer to as \emph{expert specialization training} for upcycling. Briefly, this method allocates a portion of our computational budget to independently pre-train the dense model on each of \(n\) distinct datasets, each characterized by different distributions \(\mathcal{D}_1, \ldots, \mathcal{D}_n\). This process yields \(n\) diverse and more specialized model checkpoints. We anticipated that initializing the MoE weights from these specialized checkpoints would promote expert diversification, resulting in a performance improvement.

Our experiments were conducted using dense checkpoints that contain 1.3B parameters, initially pre-trained from scratch for 1T tokens on a mixed corpus of Chinese texts, English texts, and code. We refer to this initial model as \(M_{\textrm{base}}\). Subsequently, we continued to pre-train \(M_{\textrm{base}}\) separately on an additional 100B tokens of exclusively Chinese, English, and code data, updating only the FFN part of \(M_{\textrm{base}}\). The resulting models are designated as \(M_{\textrm{cn}}\), \(M_{\textrm{en}}\), and \(M_{\textrm{code}}\), respectively. In our experiments, 
to initialize an MoE model with 8 experts, we utilized three copies of \(M_{\textrm{cn}}\), three copies of \(M_{\textrm{en}}\), one copy of \(M_{\textrm{code}}\), and one copy of \(M_{\textrm{base}}\). This setup was compared against a baseline method, which involves initializing from eight copies of \(M_{\textrm{base}}\).

The experimental results, as shown in Figure \ref{fig:multi_init}, reveal that while expert specialization training does offer a slight advantage over the baseline upcycling approach, the advantage diminishes as training progresses. By the end of 90 billion tokens of training, the difference in loss between the specialization training and the baseline is below $0.01$. We consider this difference to be marginal and not justifying the additional effort\footnote{We have trained each of \(M_{\textrm{cn}}\), \(M_{\textrm{en}}\), and \(M_{\textrm{code}}\) for 100B tokens, which altogether is roughly equivalent to 150B training of the MoE model in terms of GPU hours invested.} involved.

\end{document}